# Tuning Traditional Language Processing Approaches for Pashto Text Classification

[1]Jawid Ahmad Baktash, [2]Mursal Dawodi,  Mohammad Zarif Joya,  Nematullah Hassanzada
[1,2]LIA, Avignon University
[1,2]Avignon, France
[1]jawid.baktash1989@gmail.com, [2]mursal.dawodi@gmail.com

*Abstract*

*Today text classification becomes critical task for concerned individuals for numerous purposes. Hence, several researches have been conducted to develop automatic text classification for national and international languages. However, the need for an automatic text categorization system for local languages is felt. The main aim of this study is to establish a Pashto automatic text classification system. In order to pursue this work, we built a Pashto corpus which is a collection of Pashto documents due to the unavailability of public datasets of Pashto text documents. Besides, this study compares several models containing both statistical and neural network machine learning techniques including Multilayer Perceptron (MLP), Support Vector Machine (SVM), K Nearest Neighbor (KNN), decision tree, gaussian naïve Bayes, multinomial naïve Bayes, random forest, and logistic regression to discover the most effective approach. Moreover, this investigation evaluates two different feature extraction methods including unigram, and Time Frequency Inverse Document Frequency (IFIDF). Subsequently, this research obtained average testing accuracy rate 94% using MLP classification algorithm and TFIDF feature extraction method in this context.*

## 1. Introduction

The evolution of technology instigated existence of overwhelming number of electronic documents therefore text mining becomes a crucial task. Many businesses and individuals use machine learning techniques to accurately and quickly classify documents. On the other hand, more than 80% of organization information is in electronic format including news, email, data about user, reports, etc. [16]. Text mining attracted the attention of researchers to automatically figure out the patterns of millions electronic texts. Among other opportunities, this provides facility for users to discover the most desirable text/document.

Today, the topic of text labeling in the field of text mining and analysis is a momentous and state of the art research [4]. In the last decade, automatically assigning text documents to predefined classes has possessed the consideration of text mining experts. Some of the prior studies used machine learning techniques that utilize a set of pre-trained labeled texts for learning how to classify unseen texts [4]. Text and document classification have variety of application areas including filtering texts and documents e.g. emails, automatic question-answer systems, text and document classification, and any application that handles texts or documents. Despite, multiple works contributed in this regard few attempts has been done in Pasto natural language processing.

Pashto is a resource-poor language and the unavailability of a standard, public, free of cost datasets of text documents is a major obstacle for Pashto document classification. Automatic text document classification and comparatively analyze the performance of different models are the main gaps in Pashto text mining. This research is the first attempt to classify Pashto document into eight classes including Sport, History, Health, Scientific, Cultural, Economic, Political, and Technology. Finally, this research fills these gaps by:
- Designing a Pashto document dataset and make it publicly and free of cost available in the future.
- Classifying the designed datasets using 16 different models (combination of either Time Frequency Inverse Document Frequency (TFIDF) or Unigram with MLP, Multinomial Naïve Bayes, Gaussian Naïve Bayes, Random Forest, Logistic Regression, K Nearest Neighbor (KNN), and Decision Tree)
- Comparing the result of different models

The next section discusses previous information on the Pashto language and the ongoing work in this area is given in Section 3. Section 4 describes the proposed methods for this work. Subsequently, a description of the models used is described in Section 5. Section 6 explains the experiments and evaluation methods used in this study. Similarly, Section 7 contains the results and discussion of this study. Finally, Section 8 concludes this article.

## 2. Pashto Language

Pashto is an Iranian language, a branch of the Indo-European language family, spoken natively by a majority of Afghans, more than seven million Pakistani, and 5000 Iranian [19]. Pashtuns, people whose mother tongue is Pashto, usually live in the south of Afghanistan and north of Pakistan. This language has three main distinct dialects based on the geographic location of native Pashtun residents. The diversity of dialects even effects on spelling of Pashto text since some speakers pronounce the "sh" like "x" in Greek or "ch" in Germany rather than "sh" in English [19]. Besides, no transliteration standard exists for rendering the Pashto text to the roman alphabet and that is why both Pashto and Pashtu are the correct spelling form [19]. However, one can find some official recommendations relevant to Pashto writing and speaking. Moreover, it does not have any standard rules for writing and pronunciation therefore the authors often write one word in several ways and the speakers pronounce them in various ways [19]. The representation of letters in this language is similar to Arabic and Persian with some extra characters. Fig 1 demonstrates the alphabet representation in the Pashto language.

Pashto differentiate nouns based on genders and distinguishes the form of verbs and pronouns for masculine and feminine nouns, as an example, دا د هغې مور ده (daa de haghe mor da) means she is her mother and دا د هغه پلار دی (daa de hagha pelar de) indicates he is his father. Morphemes like plural morphemes in Pashto added another challenge to this language [10], e.g. the plural form of زوی (son) is زامن (zaamen, sons) while کتابونه (ketaboona, books) is the plural form of کتاب (ketab, book) and the plural form of انجلی ( enjeley, girl) is انجونی (anjoone, girls). Besides, news, articles, and other online and offline scripts are not written/typed by Pashto native speakers hence the probability of grammar and spelling error is high [19]. Additionally, grammar in Pashto is not as traditional as other Indo-European languages. Although nowadays several Pashto grammar books are published. Still, they have contradicted each other in some parts [19]. Furthermore, other languages spoken in the vicinity of Pashtun areas have major influences on this language that caused arriving of foreign words in Pashto for instance. some Pashtuns combine Urdu or Dari words with Pashto while speaking or in their written text.

## 3. Related Works

Many studies on document classification have already been conducted on international and western languages. Rasjid Z.E. and Setiawan R. [18] compared the performance of KNN and Naïve Bayes classification methods. They claimed that KNN with 55.17% F1-measure rate had more impressive results than Naïve Bayes (with F1-measure=39.01%) in document classification problems. As a most recent works in text document classification, Gutiérrez et al. [8] developed a COVID 19 document classification system. They compared several algorithms including: SVM, LSTM, LSTMreg, Logistic Regression, XML-CNN, KimCNN, Bertbase, Bertlarg, Longformer, and BioBert. The best performance achieved by BioBert with an accuracy of 75.2% and micro-F1 score of 0.862 on the test set. Similarly, Dadgar S.M.H. et al. [3] obtained impressive results exploiting TFIDF and SVM methods to classify news documents. They obtained 97.84% and 94.93% Precisions after evaluating their experiments on BBC and 20NewsGroups datasets.

In recent years some researchers started to work on document classification on Asian and local languages. As an example, Mohtashami and Bazrafkan [13] developed a platform to automatically categorize Persian documents. They feed their model using texts from Persian news. They considered seven classes including Social, Economic, Medical, Political, Cultural, Art, and Sport. They produced more than 100 texts, divided into 80 learnings and 20 tests for each group. The KNN with LTC feature generation outperformed in this context by obtaining 80% accuracy. Similarly, Farid D.Md. et al. [6] proposed two hybrid techniques using Decision Tree and Naïve Bayes algorithm. They tested their proposed methods on 10 real benchmark datasets. Final experiment results expressed that both proposed hybrid methods generated impressive results in the real-life classification problems. Later, Ghasemi and Jadidinejad [7] used character level convolutional neural network to classify Persian documents. They obtained 49% accuracy which was much higher compared to the results of Naïve Bayes and SVM.

Hakim et al. [9] used TFIDF to develop an automatic document classification for grouping news articles in Bahasa Indonesia. They obtained impressive result 98.3 % accuracy. Similarly, Trstenjak et al. [20] exploits KNN and TFIDF for classifying documents into four classes including sport, finance, daily news, and politics. The overall successful classifications were 92%,78%, 65%, and 90% for each particular category, respectively. On the other hand, the work by Lilleberg [11] suggests that using a combination of TFIDF and W2Vec algorithms when document clustering is better than both single methods. The main reason is that the hybrid algorithm can record more features, for example, W2Vec can capture semantic features that are not possible with TFIDF alone.

Şahİn G. [22] classified Turkish documents using Word2Vec to extract vectors of words and SVM for classification. They obtain the average F-measure score 0.92 for seven distinct categories including art and culture, economics, technology, sport, magazine, politics, and health. Subsequently, Baygin M. [1] used Naïve Bayes method and n-gram features to classify documents in Turkey into economic, health, sports, political and magazine news groups. They performed their proposed model on 1150 documents written in Turkey. The best performance achieved by 3-gram technique with 97% accuracy on sport, politics, and health documents, 98% on magazine, and 94% on economic documents.

Similarly, Pervez et al. [15] obtained impressive results using single layer convolutional neural network with different kernel sizes to classify Urdu documents. They evaluated the model on three different Urdu datasets including NPUU, naïve, and COUNTER. NPUU corpus consists of sport, economic, environment, business, crime, politics, and science and technology Urdu documents. Likewise, naïve contains Urdu document related to sports, politics, entertainment, and economic. Finally, the main document classes in COUNTER dataset are business, showbiz, sports, foreign, and national. Consequently, they obtained 95.1%, 91.4%, and 90.1% accuracy on naïve, COUNTER, and NPUU datasets, respectively.

Pal et. al [14] categorized Indi poem documents into three classes romance, heroic, and pity according to the purpose of the poem. They evaluated several machine learning techniques. The maximum accuracy 56%, 54%, 44%, 64%, and 52% using Random Forest, KNN, Decision Tree, Multinomial Naïve Bayes, SVM, and Gausian Naïve Bayes. For the first time, Rakholia [17] classified Gujrati documents using Naïve Bayes method with and without feature selection which obtained 75.74% and 88.96% accuracy respectively.

As on date, there is no document classifier available for Pashto language. The only work on Pashto text classification that is in some points related to our work was conducted by S. Zahoor et al. [21] has developed an optical character recognition system that captures images of Pashto letters and automatically classifies them by predicting a single character.

## 4. Method

This section details the methods followed to accomplish the study for categorizing Pashto sentences and documents. Figure 2. represents the main processes involved in this classification study.

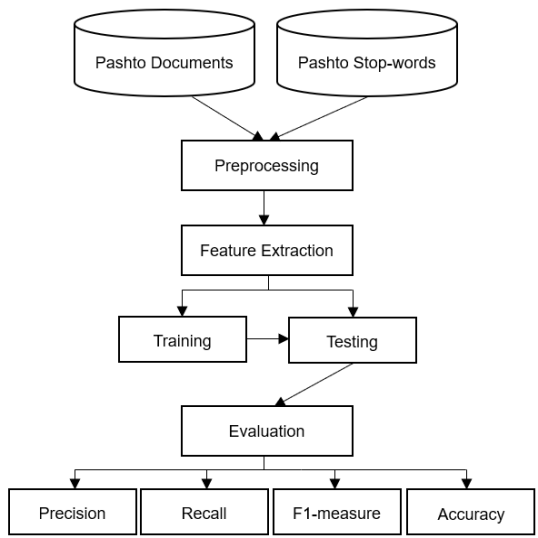

Figure 1. Phases involved in Automatic Text Classification

### 4.1 Corpora Preparation and Preprocessing

In order to make corpus for text document classification analysis, this research gathered 800 manuscripts from several online books, articles, and webpages. Consequently, manually labeled and set a number for them in relevance to the related category. We collected 100 Pashto documents for each class including history, technology, sport, cultural, economic, health, politic, and scientific.

In the preprocessing step we applied some spelling and grammar modification. In addition, we removed any noisy and senseless symbols including non-language characters, special symbols, numeric values, and URLs. Later, we applied tokenization to split each sentence to lexicons/tokens (e.g. ['واه', 'ډیر', 'ښکلی']) using Hazem library. Procurement root of separate tokens in Pashto is a more challenging task due to morphemes and other issues in Pashto literature. Thus, this work used lexicons in their default forms. As a result, we standardized and normalized the texts within the documents.

## 4.2 Feature Extraction and Selection

This research applied two common algorithms in text mining field [9], TFIDF and unigram to extract features from Pashto text and create vector representation of each document. Bag of word (BOW) technique represents the entire words as the bags regardless of their order or grammar. Unigram is a popular BOW method which considers each single word as a bag. In the next step we select the most significant attributes from extracted features therefore this study exploits Chi23 feature selection method to accomplish this propose.

### 4.2.1 Term Frequency Inverse Document Frequency

TFIDF calculates the weight of each word in the created lexicon. It consists of two functions term frequency (tf) and invers document frequency (idf). The term frequency is related to how many times a particular word appears within a document or text where inverse document frequency is the calculation from the log of the inverse probability of that word occurrences in all documents [9]. Given a dataset, $D = \{d_1, d_2, \ldots, d_n\}$ where each document contains $d_i = \{w_1, w_2, \ldots, w_n\}$ the TFIDF is defined as equation 1.

$$TFIDF(w, d, D) = tf(w, d) * idf(w, D) \quad (1)$$

Assume that we want to calculate the weight of term "خیر" from Pashto short sentences which are represented in Table 1. This word only appears once in first and twice in the second texts. Therefore, TFIDF of this word is calculating as below:

$$TFIDF(\text{خیر}, text_1) = 1 * \log\left(\frac{3}{2}\right) = 0.23$$

$$TFIDF(\text{خیر}, text_2) = 2 * \log\left(\frac{3}{2}\right) = 0.46$$

Table 1 Example of Pashto text

| Seq | Pashto text |
|---|---|
| 1 | سهار مو په خیر |
| 2 | خیر دی. ورځ مو په خیر |
| 3 | ښه قسمت درته غوارم |

## 4.3 Classification Methods

As mentioned in prior sections, this work observed 8 different classifiers methods including Naïve Bayes, Multinomial Naïve Bayes, Nearest Neighbor, Random Forest, Decision Tree, SVM, MLP, and Logistic Regression. We evaluated different sets of training and testing corpuses for Pashto documents classification project with 20/80, 30/70, 40/60 and 50/50 ratio. In the final step, we allocated 80% of data records for training and 20% for testing proposes. In order to evaluate performance of separate methods, we used accuracy, Precision, Recall and F1-measure approaches.

**Naïve Bayes**

Naïve Bayes is a statistical classification based on the Bayes theorem that is most useful with small datasets. However, it is easily scalable with large corpora. It calculates the probability of two events occurring based on the probability of each event occurring separately. Here, each vector representing a text contains information about the probability of each term occurring in the given text.

The likelihood of a text is taken from the probability of words within that text and the probability of the text appearing with the same length. Whenever the probability of betting on a particular attribute is zero, Naïve Bayes is unable to make a valid prediction.

Assume a dataset, $D = \{R_1, R_2, \ldots, R_n\}$ and each data record as $R_i = \{r_1, r_2, \ldots, r_n\}$. The dataset contains a set of classes $C = \{c_1, c_2, \ldots, c_n\}$. Therefore, each record within the training dataset has a certain class label. Naïve Bayes predicts to which class an experimental sample depends through discovering the class with the uppermost posterior probability, conditioned on R. The equation 2 represents the Naïve Bayes theorem where $P(C_i|R)$ indicates class posterior probability, $P(R|C_i)$ represents the likelihood of a word x in that class, P(c) shows class prior probability, and P(x) means predictor prior probability.

$$P(C_i|R) = \frac{P(R|C_i)P(C_i)}{P(R)} \quad (2)$$

In this research, we used wo common types of Naïve Bayes which are Gaussian Naïve Bayes and Multinomial Naïve Bayes. In multinomial Naïve Bayes, attribute vectors contain elements that determine the frequency of occurrence of a certain attribute. In contrast to Gaussian Naïve Bayes, which is effective in more general classifications problems, this method is useful when samples are taken from a standard shared dataset.

### Support vector machine

The SVM is a supervised classification technique and as its name implies it represents the training data as support vectors. Similar to Naïve Bayes, SVM can provide accurate results even with small datasets [2]. SVM divides the space between vectors belong to and vectors not related to a tag. In other words, SVM specifies a hyperplane between the positive and negative examples of the training set. There is a margin between the hyperplane and the nearest positive and negative sample. This method is memory-efficient because it effectively manages high-dimensional spaces by using only a subset of training data to estimate probabilities. However, it makes predictions using five-fold cross-validation, which is very expensive.

### Decision Tree

The decision tree is a simple supervised learning algorithm that is widely used in regression and classification problems. This algorithm learns the rules of decision making while performing training tasks from training data. In addition, it uses the tree display model for making predictions where the inner node represents a particular feature and the leaves are class labels. This method performs binary splitting of data on all leaves. The decision tree requires a small amount of data in which the data can be categorical and numerical. However, it may generate more complex and not generalizable trees. In addition, this is an unsTable algorithm because a small number of changes in data create a completely different tree.

### Random Forest

The random forest addresses the problem of overfitting in the decision tree algorithm by creating multiple trees and assigning objects to the class with the highest number of votes obtained from all trees. The main weak point of this algorithm is its complexity.

### Logistic regression

Logistic regression uses a logistic function to create a predictive model. It is useful for handling a dichotomous corpus where binary classes exist. The logistic regression can be binomial, multinomial, and ordinal based on the total number of possible variable types.

### K Nearest Neighbor

KNN is a popular classification method that classifies unseen data based on similarity measurements to the k most similar records in the training/testing dataset [5,12]. This project fine-tuned the value of K finally the result shows that the optimum value for k occurs at k=5. In the first step, it loads the data model, then measures the similarity between the target item and other training data items by distance. It then sorts the result from the closest distance to the furthest distance. Finally, it predicts the corresponding class based on the value of K.

## Multilayer Perceptron

MLP is a neural network classifier which is a subset of machine learning consists of neurons and layers [Fetch 2019]. Neural Networks usually learn like human learning mechanism. A neural network entails graphs of mathematical modeled neurons connected with particular weights. Whenever, the NN model has more than one hidden layer it implies deep neural network. MLP consists of several layers including one input, one output, and hidden layers as represented in Figure 3.

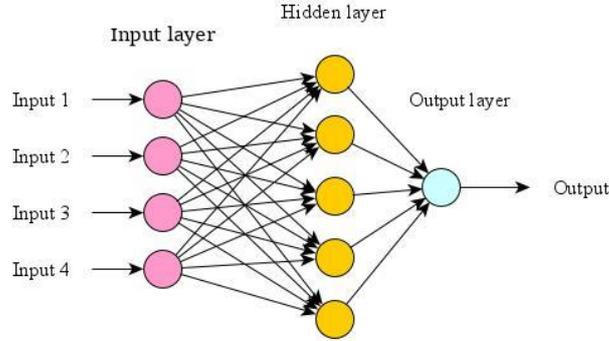

Figure 3. Architecture of MLP with one hidden layer

The outputs of one perceptron are fed as input to subsequent perceptron. This experiment used single hidden layer with 20 neurons. We used back propagation and gradient descent to provide ability to propagate errors back to earlier layers. Moreover, we shuffled the samples to reduce noise by feeding different inputs to neurons in each iteration and as a result make good generalization. The activation function used in this model is Rectified Linear Unit (ReLU) function, which is a non-linear activation function, to decrease the chance of vanishing gradient. ReLU is defined in equation 3 where f(a) is considered to be zero for all negative numbers of a. Finally, we used Adam as optimization algorithm.

$$f(a) = \max(0, a) \text{ where } a = Wx + b \quad (3)$$

## 5. Experiments

This study evaluates the performance of eight different classifiers using separate feature extraction methods. We considered four metrics Precision (4), Recall (5), F-measure (6), and accuracy (7) to analyze the outcome of the 16 different models used in this experiment. Precision, which is called positive predictive values, is the percentage of examples that the classifier predicts accurately from the total samples predicted for a given tag. On the other hand, Recall which is also referred to as sensitivity determines the percentage of samples that the classifier predicts for a given label from the total number of samples that should be predicted for that label. Accuracy represents the performance of the model while is referred to the percentage of texts that are predicted with the correct label. We used F1 score to measure the average between Precision and Recall values. There are mainly four actual classes true real positive (TP), false real positive (FP), true real negative (TN), and false real negative (FN). TP and TN are the accurate predictions while FP and FN are related to imprecise estimations:

$$\text{True class} = \{TP_1, TP_2, \ldots, TP_n, \} \cup \{TP_1, TP_2, \ldots, TP_n, \}$$
$$\text{False class} = \{FP_1, FP_2, \ldots, FP_n, \} \cup \{FP_1, FP_2, \ldots, FP_n, \}$$

Consequently, this study computes weighted average and macro average values for Precision, Recall, and F1-measure of all classes to compare the efficiency of each technique.

$$Precision = \frac{TP_i}{TP_i + FP_i} \quad (4)$$
$$Recall = \frac{TP_i}{TP_i + FN_i} \quad (5)$$
$$F1 - measure = \frac{2PR}{P + R} \quad (6)$$
$$Accuracy = \frac{TP_i + TN_i}{TP_i + FP_i + TN_i + FN_i} \quad (7)$$

## 6. Results and Discussion

MLP with unigram feature extraction technique illustrated the best performances among others with the gained average accuracy of 94%. Besides, it obtained 0.94 as weighted average Precision, Recall, and F-measure scores. As one can see in Figure 7, maximum weighted average Precision using MLP and unigram is 0.91. The obtained results are presented in Table 2 and Figure 5 to 7. Figure 4 comparatively demonstrates obtained accuracy using different techniques. Similarly, Figure 5 comparatively represents the testing accuracy for all 16 separate approaches. Figure 5 to 7 denote the Recall, F1-measure, and Precision values, both macro average, and weighted average obtained when testing distinct techniques.

Multinomial Naïve Bayes with Unigram achieved 88% accuracy while it decreased by 7% replacing Unigram with TFIDF which indicates that it performed better with Unigram text embedding technique. However, Gaussian Naïve Bayes obtained 87% accuracy using TFIDF vector representations which is 11% higher compared to Gaussian Naïve Bayes +Unigram. Even though, Gaussian Naïve Bayes has impressive result 0.85 as weighted Precision result, but it obtained f1 score of only 0.77 due to its low Recall score of 0.76. In contrast to Gaussian Naïve Bayes, Decision Tree obtain 5% more accuracy using Unigram rather than TFIDF. Performance of Logistic Regression, SVM, Random Forest, and KNN with both TFIDF and Unigram are comparable with only 1% change in accuracy and 0-0.2 variation in f1 scores.

The combination of SVM and unigram represented 84% average accuracy while this value is reduced by 1% using TFIDF. Therefore, similar to several classification studies SVM performed good in Pashto text document classification. In contrast to the work by Mohtashami and Bazrafkan [13], KNN attained only 71% as average accuracy using TFIDF method that is decreased to 70% after altering the feature extraction method from TFIDF to Unigram. The least performance belongs to Decision Tree method with TFIDF technique in this comparison experiment which is only 64% accuracy. This method also has low performance (with F1-measure of 0.69) using unigram extraction method.

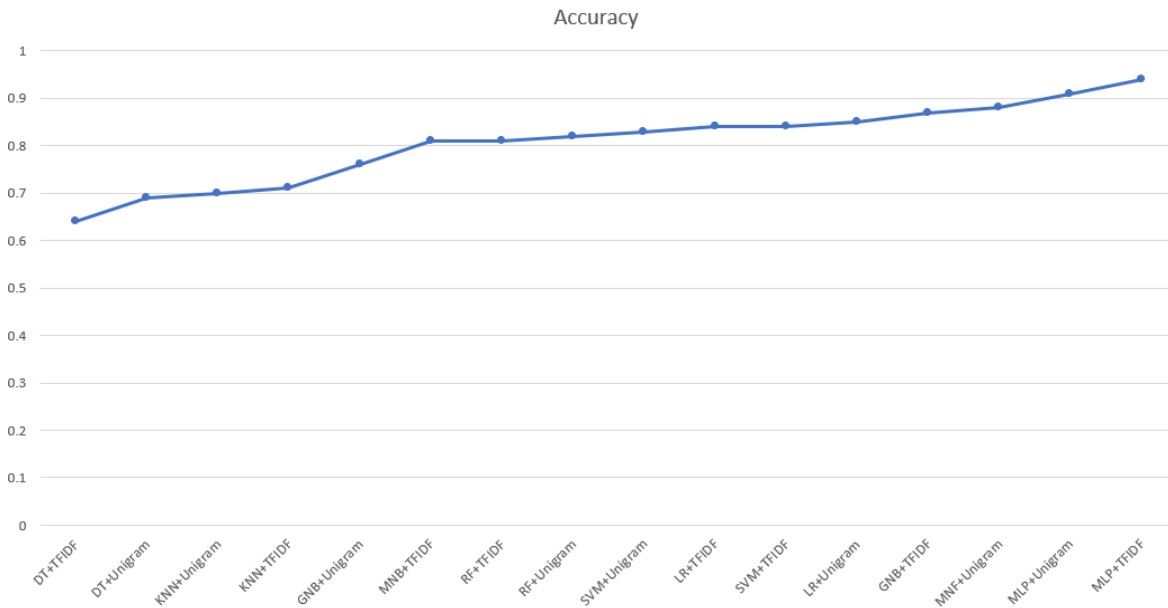

Figure 4. Comparison of obtained accuracy

Currently, this research is the only study on in the field of Pashto document classification. Mohtashami and Bazrafkan [13] conducted similar research on Persian document classification. In their study, KNN had the best performance with 90% accuracy. All in all, our model gives 3% less testing accuracy compared to [13].

Table 2. Average accuracy using different classification and feature extraction techniques

| Technique | Feature Extraction Method | Accuracy |
|---|---|---|
| Gaussian Naïve Bayes | Unigram | 0.76 |
|  | TFIDF | 0.87 |
| Multinomial Naïve Bayes | Unigram | 0.88 |
|  | TFIDF | 0.81 |
| Decision Tree | Unigram | 0.69 |
|  | TFIDF | 0.64 |
| Random Forest | Unigram | 0.82 |
|  | TFIDF | 0.81 |
| Logistic Regression | Unigram | 0.85 |
|  | TFIDF | 0.84 |
| SVM | Unigram | 0.83 |
|  | TFIDF | 0.84 |
| K Nearest Neighbor | Unigram | 0.7 |
|  | TFIDF | 0.71 |
| Multilayer Perceptron | Unigram | 0.91 |
|  | TFIDF | 0.94 |

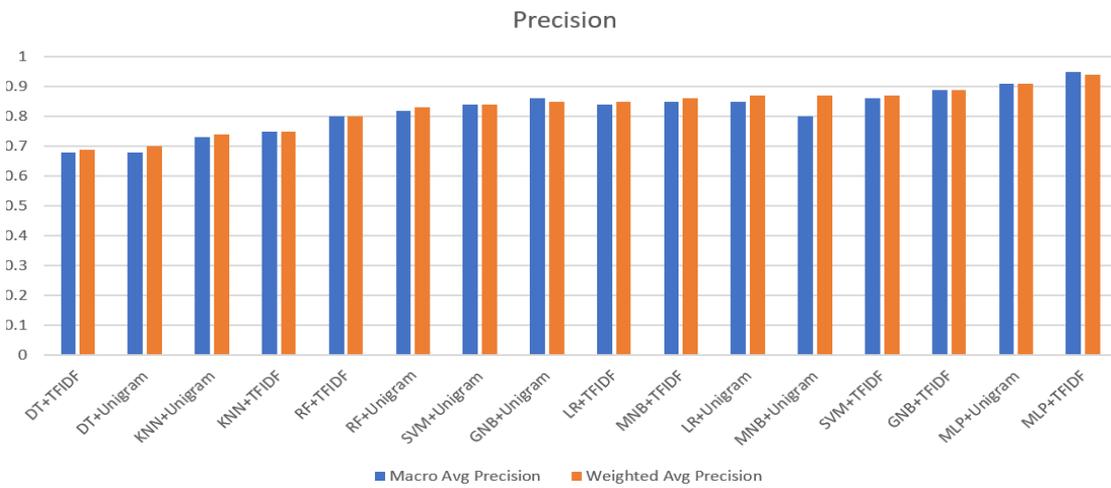

Figure 5 Macro average Precision and weighted average Precision

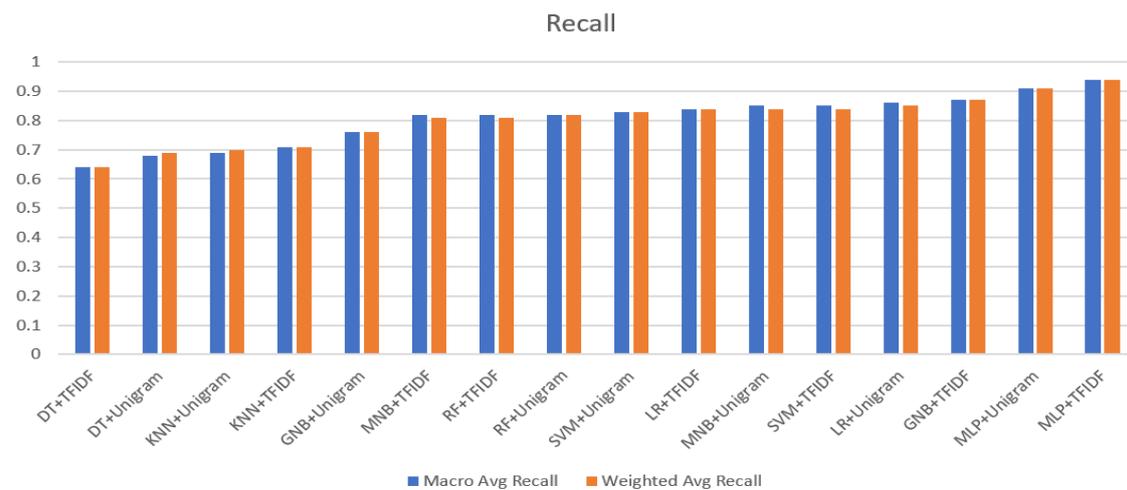

Figure 6 Macro average Recall and Weighted Average Recall

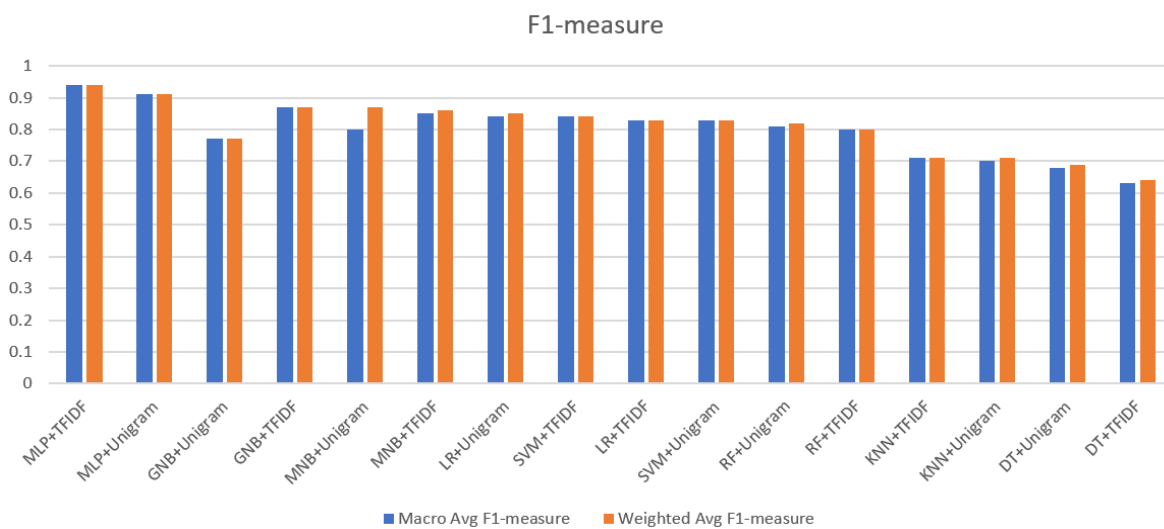

Figure 7 Macro average and weighted average F1-measure

The outcome of each model is different according to the separate class label as illustrated in Tables 3-11. As an example, KNN employed unigram has 0.98 f1 score related to History tag. However, it obtained only 0.37 for scientific documents. Similarly, all models illustrated good f1-score for documents relevant to History except Gaussian Naïve Bayes. DT achieved high f1 score only by predicting documents related to History. Experiments show that MLP models and the combined model of Random Forest with Unigram more accurately predicts cultural documents compared to other models. MLP with TFIDF and Gaussian Naïve Bayes with Unigram with 0.95 and 0.93 f1 score have the most accurate Economic class predictions in this experiment. On the other hand, the implementation of Gaussian Naïve Bayes and SVM with Unigram represents the most precise results in the context of health documents. Similarly, MLP with TFIDF obtains the highest f1 score of 1 on predicting Politic documents. All models failed to predict scientific documents precisely except MLP + TFIDF model with f1 score of 0.89. MLP with Unigram with f1 score 0.98 best performed in discovering texts related to Sport class. Similarly, Random Forest and Gaussian Naïve Bayes with TFIDF came in second with f1 score 0.95 in this era. Likewise, MLP+TFIDF and Gaussian Naïve Bayes + TFIDF models best predicts texts belonging to Technology class with f1 score 1 and 0.97 respectively.

This study has some limitations due to the immature context of the Pashto language. There is not any special toolkit for processing Pashto language like Hazm for Persian language and NLTK for English language. The data set used in this study is very short with only 800 records. Additionally, this experiment only took into account 8 separate classes for Pashto documents. However, our future goal is to expand the corpus and use more hybrid algorithms to achieve better performance.

Table 3. Performance of KNN on separate tags

| KNN | | History | Culture | Economic | Health | Politic | Scientific | Sport | Technology |
|---|---|---|---|---|---|---|---|---|---|
| Unigram | Precision | 0.96 | 0.92 | 0.48 | 0.7 | 0.93 | 0.32 | 0.82 | 0.71 |
| | Recall | 1 | 0.61 | 0.62 | 0.7 | 0.74 | 0.44 | 0.64 | 0.77 |
| | F1-measure | 0.98 | 0.73 | 0.54 | 0.7 | 0.82 | 0.37 | 0.72 | 0.74 |
| TFIDF | Precision | 0.84 | 0.64 | 0.59 | 0.59 | 0.8 | 0.5 | 0.81 | 0.92 |
| | Recall | 0.94 | 0.75 | 0.81 | 0.81 | 0.57 | 0.68 | 0.71 | 0.48 |
| | F1-measure | 0.89 | 0.69 | 0.69 | 0.75 | 0.67 | 0.58 | 0.76 | 0.63 |

Table 4. Performance of SVM on separate tags

| SVM | | History | Culture | Economic | Health | Politic | Scientific | Sport | Technology |
|---|---|---|---|---|---|---|---|---|---|
| Unigram | Precision | 1 | 0.74 | 0.71 | 0.94 | 0.88 | 0.74 | 0.9 | 0.79 |
| | Recall | 0.85 | 0.85 | 0.71 | 0.88 | 0.95 | 0.82 | 0.86 | 0.71 |
| | F1-measure | 0.92 | 0.79 | 0.71 | 0.91 | 0.91 | 0.78 | 0.88 | 0.75 |
| TFIDF | Precision | 1 | 0.6 | 0.93 | 0.95 | 0.9 | 0.6 | 1 | 0.86 |
| | Recall | 1 | 1 | 0.87 | 0.72 | 0.86 | 0.75 | 0.81 | 0.8 |
| | F1-measure | 1 | 0.75 | 0.9 | 0.82 | 0.88 | 0.67 | 0.9 | 0.83 |

Table 5. Performance of Random Forest on separate tags

| RF | | History | Culture | Economic | Health | Politic | Scientific | Sport | Technology |
|---|---|---|---|---|---|---|---|---|---|
| Unigram | Precision | 1 | 0.95 | 0.8 | 0.79 | 0.68 | 0.58 | 0.85 | 0.9 |
| | Recall | 1 | 0.83 | 0.71 | 0.88 | 0.94 | 0.41 | 0.89 | 0.86 |
| | F1-measure | 1 | 0.88 | 0.75 | 0.84 | 0.79 | 0.48 | 0.87 | 0.88 |
| TFIDF | Precision | 1 | 0.83 | 0.65 | 0.83 | 0.71 | 0.61 | 0.9 | 0.83 |
| | Recall | 0.96 | 0.79 | 0.65 | 0.86 | 1 | 0.48 | 1 | 0.79 |
| | F1-measure | 0.98 | 0.81 | 0.65 | 0.85 | 0.83 | 0.54 | 0.95 | 0.81 |

Table 6. Performance of Decision Tree on separate tags

| DT | | History | Culture | Economic | Health | Politic | Scientific | Sport | Technology |
|---|---|---|---|---|---|---|---|---|---|
| Unigram | Precision | 1 | 0.72 | 0.45 | 0.58 | 0.73 | 0.59 | 0.82 | 0.56 |
| | Recall | 0.92 | 0.72 | 0.5 | 0.78 | 0.76 | 0.5 | 0.64 | 0.6 |
| | F1-measure | 0.96 | 0.72 | 0.48 | 0.67 | 0.74 | 0.54 | 0.72 | 0.58 |
| TFIDF | Precision | 1 | 0.75 | 0.42 | 0. 45 | 0.69 | 0.44 | 1 | 0.71 |
| | Recall | 0.95 | 0.63 | 0.52 | 0.82 | 0.82 | 0.41 | 0.4 | 0.52 |
| | F1-measure | 0.98 | 0.69 | 0.57 | 0.58 | 0.75 | 0.42 | 0.57 | 0.6 |

Table 7. Performance of MLP on separate tags

| MLP | | History | Culture | Economic | Health | Politic | Scientific | Sport | Technology |
|---|---|---|---|---|---|---|---|---|---|
| Unigram | Precision | 1 | 1 | 0.90 | 0.87 | 0.95 | 0.68 | 0.95 | 0.90 |
| | Recall | 1 | 0.94 | 0.82 | 0.91 | 0.91 | 0.81 | 1 | 0.86 |
| | F1-measure | 1 | 0.97 | 0.86 | 0.89 | 0.93 | 0.74 | 0.98 | 0.88 |
| TFIDF | Precision | 0.96 | 0.91 | 1 | 0.89 | 1 | 0.86 | 0.95 | 1 |
| | Recall | 0.96 | 1 | 0.90 | 0.84 | 1 | 0.92 | 0.91 | 1 |
| | F1-measure | 0.96 | 0.95 | 0.95 | 0.86 | 1 | 0.89 | 0.93 | 1 |

Table 8. Performance of Logistic Regression on separate tags

| LR | | History | Culture | Economic | Health | Politic | Scientific | Sport | Technology |
|---|---|---|---|---|---|---|---|---|---|
| Unigram | Precision | 1 | 0.89 | 0.96 | 0.8 | 0.67 | 0.59 | 1 | 0.86 |
| | Recall | 1 | 1 | 0.88 | 0.83 | 0.88 | 0.67 | 0.72 | 0.86 |
| | F1-measure | 1 | 0.94 | 0.92 | 0.82 | 0.76 | 0.62 | 0.84 | 0.86 |
| TFIDF | Precision | 0.95 | 0.8 | 0.85 | 0.76 | 0.77 | 0.87 | 0.71 | 1 |
| | Recall | 1 | 0.94 | 0.88 | 0.76 | 1 | 0.62 | 0.83 | 0.72 |
| | F1-measure | 0.97 | 0.88 | 0.86 | 0.76 | 0.87 | 0.72 | 0.77 | 0.84 |

Table 9. Performance of Gaussian Naïve Bayes on separate tags

| GNB | | History | Culture | Economic | Health | Politic | Scientific | Sport | Technology |
|---|---|---|---|---|---|---|---|---|---|
| Unigram | Precision | 0.44 | 1 | 0.93 | 0.92 | 0.79 | 0.77 | 1 | 1 |
| | Recall | 1 | 0.63 | 0.93 | 0.92 | 0.94 | 0.5 | 0.56 | 0.62 |
| | F1-measure | 0.61 | 0.77 | 0.93 | 0.92 | 0.86 | 0.61 | 0.71 | 0.77 |
| TFIDF | Precision | 0.64 | 1 | 1 | 0.86 | 0.91 | 0.75 | 1 | 0.1 |
| | Recall | 0.95 | 0.81 | 0.73 | 0.95 | 0.95 | 0.71 | 0.91 | 0.94 |
| | F1-measure | 0.76 | 0.89 | 0.84 | 0.9 | 0.93 | 0.73 | 0.95 | 0.97 |

Table 10. Performance of Multinomial Naïve Bayes on separate tags

| MNB | | History | Culture | Economic | Health | Politic | Scientific | Sport | Technology |
|---|---|---|---|---|---|---|---|---|---|
| Unigram | Precision | 0.87 | 0.8 | 1 | 0.82 | 0.76 | 0.83 | 1 | 1 |
| | Recall | 0.95 | 1 | 0.81 | 0.93 | 0.94 | 0.77 | 0.88 | 0.87 |
| | F1-measure | 0.91 | 0.89 | 0.89 | 0.87 | 0.84 | 0.8 | 0.93 | 0.93 |
| TFIDF | Precision | 0.82 | 0.94 | 0.73 | 0.86 | 0.6 | 0.78 | 1 | 1 |
| | Recall | 1 | 0.71 | 0.83 | 0.86 | 1 | 0.74 | 0.7 | 0.65 |
| | F1-measure | 0.9 | 0.81 | 0.78 | 0.86 | 0.75 | 0.76 | 0.82 | 0.79 |

# 7. Conclusion

This paper is one of the first state of the art researches in Pashto literature text classification analysis. It built the first Pashto documents corpus. It also made lexicon list of Pashto words and developed multiple classification framework to categorize Pashto documents. This study obtained high accuracy with some classifiers. The highest accuracy achieved by implementing MLP with TFIDF methods with 94% accuracy. In our future work we will expand our dataset and add lemmatization task. Moreover, we will observe recurrent neural network and convolutional neural net in this context.

# 8. References

[1] M. Baygin, "Classification of Text Documents Based on Naive Bayes Using N-Gram Features," in 2018 International Conference on Artificial Intelligence and Data Processing (IDAP), Malatya, Turkey: IEEE, 2018, pp. 1-5, https://doi.org/10.1109/IDAP.2018.8620853.

[2] B.E. Boser, M.G. Isabelle, and V.N. Vladimir, "A Training Algorithm for Optimal Margin Classifiers," in Proceedings of the Fifth Annual Workshop on Computational Learning Theory, COLT '92. Pittsburgh, Pennsylvania, USA: Association for Computing Machinery, 1992, pp. 144-152, https://doi.org/10.1145/130385.130401.

[3] S.M.H. Dadgar, S.A. Mohammad, and M.M. Farahani, "A Novel Text Mining Approach Based on TF-IDF and Support Vector Machine for News Classification," in 2016 IEEE International Conference on Engineering and Technology (ICETECH), 2016, pp. 112-116, https://doi.org/10.1109/ICETECH.2016.7569223.

[4] M. Dawodi, M.Z. Joya, N. Hassanzada, J.A. Baktash, and T. Wada, "A Comparative Study of Machine Learning Methods and Feature Extraction Methods for Dari Sentiment Analysis," Information, vol. 23, no. 2, pp. 117-137, 2020.

[5] M. Dawodi, T. Wada, and J.A. Baktash, "An Intelligent Recommender System Supporting Decision-Making on Academic Major," Information: International Information Institute (Tokyo), vol. 22, no. 3, pp. 241-254, 2019.

[6] D.Md. Farid, L. Zhang, M.R. Chowdhury, M.A. Hossain, and R. Strachan, "Hybrid Decision Tree and Naïve Bayes Classifiers for Multi-Class Classification Tasks," Expert Systems with Applications, vol. 41, no. 4, pp. 1937-1946, 2014, https://doi.org/10.1016/j.eswa.2013.08.089

[7] S. Ghasemi and H.A. Jadidinejad, "Persian Text Classification via Character-Level Convolutional Neural Networks," in 2018 8th Conference of AI & Robotics and 10th RoboCup Iranopen International Symposium (IRANOPEN), Qazvin: IEEE, 2018, pp. 1-6, https://doi.org/10.1109/RIOS.2018.8406623.

[8] B.J. Gutiérrez, J. Zeng, D. Zhang, Y. Zhang, and Y. Su, "Document Classification for COVID-19 Literature," ArXiv:2006.13816 [Cs], Retrieved Sep 1, 2020 from http://arxiv.org/abs/2006.13816.

[9] A. A. Hakim, A. Erwin, K. I. Eng, M. Galinium, and W. Muliady, "Automated Document Classification for News Article in Bahasa Indonesia Based on Term Frequency Inverse Document Frequency (TF-IDF) Approach," in 2014 6th International Conference on Information Technology and Electrical Engineering (ICITEE), Yogyakarta, Indonesia, 2014, pp. 1-4

[10] U. Kamal, I. Siddiqi, H. Afzal, and A. Ur Rahman, "Pashto Sentiment Analysis Using Lexical Features," in Proc. Mediterranean Conf. on Pattern Recognition and Artificial Intelligence (MedPRAI), Tebessa, Algeria, 2016, pp. 121-124, doi: 10.1145/3038884.3038904.

[11] J. Lilleberg, Y. Zhu, and Y. Zhang, "Support Vector Machines and Word2vec for Text Classification with Semantic Features," in Proc. IEEE 14th Int. Conf. on Cognitive Informatics & Cognitive Computing (ICCI*CC), Beijing, China, 2015, pp. 136-140, doi: 10.1109/ICCI-CC.2015.7259377.

[12] M. Mohammadi, M. Dawodi, T. Wada, and N. Ahmadi, "Comparative Study of Supervised Learning Algorithms for Student Performance Prediction," in Proc. Int. Conf. on Artificial Intelligence in Information and Communication (ICAIIC), 2019, pp. 124-127, doi: 10.1109/ICAIIC.2019.8666728.

[13] E. Mohtashami and M. Bazrafkan, "The Classification of Persian Texts with Statistical Approach and Extracting Keywords and Admissible Dataset," Int. J. of Computer Applications, vol. 101, no. 5, pp. 18-20, 2014, doi: 10.5120/17683-8541.

[14] K. Pal and V. P. Biraj, "Automatic Multiclass Document Classification of Hindi Poems Using Machine Learning Techniques," in Proc. Int. Conf. for Emerging Technology (INCET), Belgaum, India, 2020, pp. 1-5, doi: 10.1109/INCET49848.2020.9154001.

[15] A. M. Pervez, J. Zheng, I. R. Naqvi, M. Abdelmajeed, A. Mehmood, and M. T. Sadiq, "Document-Level Text Classification Using Single-Layer Multisize Filters Convolutional Neural Network," IEEE Access, vol. 8, pp. 42689-42707, 2020, doi: 10.1109/ACCESS.2020.2976744.

[16] P. Raghavan, "Text Centric Structure Extraction and Exploitation (Abstract Only)," in Proc. 7th Int. Workshop on the Web and Databases: Colocated with ACM SIGMOD/PODS 2004 (WebDB '04), New York, NY, USA, 2004, doi: 10.1145/1017074.1022586.

[17] R. M. Rakholia, J. R. Saini, and Narmada College of Computer Application, "Classification of Gujarati Documents Using Naïve Bayes Classifier," Indian Journal of Science and Technology, vol. 10, no. 5, pp. 1-9, 2017, doi: 10.17485/ijst/2017/v10i5/103233.

[18] Z. E. Rasjid and S. Reina, "Performance Comparison and Optimization of Text Document Classification Using K-NN and Naïve Bayes Classification Techniques," Procedia Computer Science, vol. 116, pp. 107

[19] Tegey, H., and Robson, B., "A Reference Grammar of Pashto," ERIC, ED399825, 1996.

[20] B. Trstenjak, S. Mikac, and D. Donko, "KNN with TF-IDF Based Framework for Text Categorization," Procedia Engineering, vol. 69, pp. 1356-1364, Jan. 2014. DOI: 10.1016/j.proeng.2014.03.129.

[21] S. Zahoor, S. Naz, N. H. Khan, and M. I. Razzak, "Deep Optical Character Recognition: A Case of Pashto Language," Journal of Electronic Imaging, vol. 29, no. 2, p. 023002, Mar. 2020. DOI: 10.1117/1.JEI.29.2.023002.

[22] G. Şahİn, "Turkish Document Classification Based on Word2Vec and SVM Classifier," in 2017 25th Signal Processing and Communications Applications Conference (SIU), Antalya, Turkey, 2017, pp. 1-4. DOI: 10.1109/SIU.2017.7960552.